\pdfoutput=1
\documentclass[11pt]{article}

\usepackage[preprint]{acl}

\usepackage{times}
\usepackage{latexsym}

\usepackage[T1]{fontenc}
\usepackage[utf8]{inputenc}

\usepackage{microtype}

\usepackage{inconsolata}

\usepackage{graphicx}
\usepackage{caption}
\usepackage{subcaption}
\usepackage{amsmath}

\usepackage{enumitem}
\usepackage{caption}
\usepackage{booktabs}
\usepackage{xcolor}
\definecolor{c1}{HTML}{0c8918}
\definecolor{c2}{HTML}{dc3023}

\usepackage{stfloats}
\usepackage{pifont}
\usepackage{makecell}

\title{Towards Robustness of Text-to-Visualization Translation against Lexical and Phrasal Variability}

\author{Jinwei Lu$^{1}$, Yuanfeng Song$^{2}$, Haodi Zhang$^1$,  \\
  \textbf{Chen Zhang$^3$, Raymond Chi-Wing Wong$^4$} \\
  $^1$Shenzhen University, Shenzhen, China $^2$AI Group, WeBank Co., Ltd, Shenzhen, China \\
  $^3$ The Hong Kong Polytechnic University, Hong Kong SAR, China \\
  $^4$ The Hong Kong University of Science and Technology, Hong Kong SAR, China\\
\\}

\begin{document}
\maketitle
\begin{abstract}
Text-to-Vis is an emerging task in the natural language processing (NLP) area that aims to automatically generate data visualizations from natural language questions (NLQs). Despite their progress, existing text-to-vis models often heavily rely on lexical matching between words in the questions and tokens in data schemas. This overreliance on lexical matching may lead to a diminished level of model robustness against input variations.
In this study, we thoroughly examine the robustness of current text-to-vis models, an area that has not previously been explored. In particular, we construct the first robustness dataset \textbf{nvBench-Rob}, which contains diverse lexical and phrasal variations based on the original text-to-vis benchmark nvBench. Then, we found that the performance of existing text-to-vis models on this new dataset dramatically drops, implying that these methods exhibit inadequate robustness overall. Finally, we propose a novel framework based on Retrieval-Augmented Generation (RAG) technique, named \textbf{GRED}, specifically designed to address input perturbations in these two variants. The framework consists of three parts: NLQ-Retrieval \underline{G}enerator, Visualization Query-Retrieval \underline{Re}tuner and Annotation-based \underline{D}ebugger, which are used to tackle the challenges posed by natural language variants, programming style differences and data schema variants, respectively. Extensive experimental evaluations show that, compared to the state-of-the-art model RGVisNet in the Text-to-Vis field, GRED performs better in terms of model robustness, with a 32\% increase in accuracy on the proposed nvBench-Rob dataset.

\end{abstract}

\section{Introduction}
\begin{figure*}[h!]
    \centering
    \includegraphics[width=\textwidth]{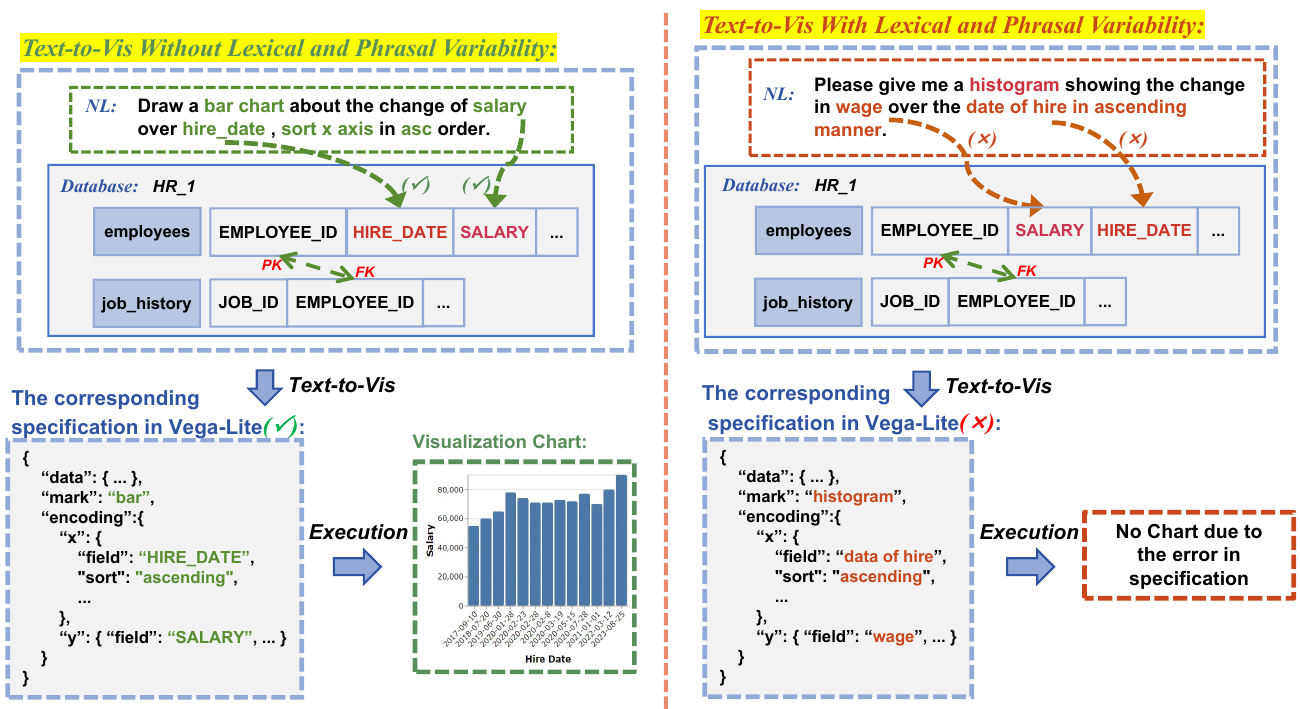}
    %\vspace{-5pt}
    \caption{(a) Text-to-vis is dedicated to converting natural language questions (NLQs) into data visualizations (DVs). The current approach heavily relies on explicit matching between words within the NLQs and the table schema. (b) The robustness of existing text-to-vis methods is limited. When small variations in NLQs and table schemas appear, the text-to-vis model fails to generate correct outputs (marked with \textcolor{red}{`$\times$'} in red color).}
    \label{fig:Text-to-Vis and Robustness}
    \vspace{-10pt}
\end{figure*}
Data visualization (DV) has emerged as an indispensable tool in the industry for extracting insights from massive data. It surpasses verbal expressions, offering a clear and effective presentation of insights derived from raw data. The process of creating DVs involves programming declarative visualization languages (DVLs) to select relevant data and determine how to present it. With a wide variety of different DVLs available—each characterized by its own distinctive grammar and syntax, such as Vega-Lite \cite{Satyanarayan2018VegaLiteAG}, ggplot2 \cite{GmezRubio2017ggplot2E}, ZQL \cite{Siddiqui2016EffortlessDE}, and ECharts \cite{Li2018EChartsAD}—the need for considerable domain knowledge and proficiency in DVL is required, posing a particularly challenge to those who lack technical expertise.\par

To enhance the accessibility of DV, a task named \textit{text-to-vis} has been proposed, which offers a mechanism to automatically transform natural language questions (NLQs) into DV charts \cite{10.1145/3448016.3457261,song2022rgvisnet}. Exemplified in Figure~\ref{fig:Text-to-Vis and Robustness}, the text-to-vis system requires users to simply ask an NLQ, such as, \textit{``Draw a bar chart about the change of salary over hire\_date, sort x axis in asc order.''} It then automatically generates the final DV, such as a bar chart, by interfacing with the database, thereby circumventing the need for users to code directly in a DVL.\par

To deploy text-to-vis models in real-life, it is crucial for these models to possess the capability to handle NLQs from diverse users. 
Therefore, the \textit{robustness} of the model plays an important role in evaluating the performance of text-to-vis models. High model performance requires robust performance on noisy inputs. 
However, the robustness of text-to-vis models poses a significant challenge. In our analysis (Section~\ref{sec:rob_analy}), we found that even small perturbations in the input may significantly reduce the performance of existing text-to-vis models. Furthermore, there is still a lack of dedicated robustness datasets and studies in the field to effectively evaluate the robustness of text-to-vis models. \par

We notice that the NLQs in the original text-to-vis dataset nvBench \cite{10.1145/3448016.3457261} usually explicitly mention the information present in the database, like explicit mentions of column names. This characteristic makes the test results of nvBench unsuitable for evaluating the robustness of the text-to-vis models. It is difficult to ascertain whether the model simply memorizes the explicitly mentioned schema, such as column names, or if it genuinely learns the natural mapping relationship between the NLQ and data schema. \par

The lack of large-scale datasets is one of the significant factors that limits the robustness studies in the text-to-vis field. In this work, we propose the first comprehensive robustness dataset named \textbf{nvBench-Rob} to evaluate the robustness of the text-to-vis models. %nvBench-Rob is based on nvBench \cite{}, a large-scale text-to-vis benchmark. 
nvBench-Rob aims to provide a comprehensive evaluation of models based on two variants: NLQ and data schema, as shown in Figure~\ref{fig:Text-to-Vis and Robustness}. With these two variants, we thoroughly examine the robustness of the current text-to-vis models, an area that has not previously been explored. We found that the performance of existing text-to-vis models dramatically drop, implying these methods exhibit inadequate robustness. 

To enhance the robustness of text-to-vis models, we propose a novel framework named \textbf{GRED} based on the Retrieval-Augmented Generation (RAG)-based technique for Large Language Models (LLMs) \cite{roziere2023code,touvron2023llama,gunasekar2023textbooks,anil2023palm,openai2024gpt4}. This framework comprises three core components: NLQ-Retrieval \underline{G}enerator, DVQ-Retrieval \underline{Re}tuner, and Annotation-based \underline{D}ebugger, aimed at addressing variants of NLQs, differences in programming styles, and changes in data schema, respectively. 
\footnote{DVQ refers to Data Visualization Query \cite{10.1145/3448016.3457261, song2022rgvisnet}, which is a widely-used intermediate representation that connects NLQ with the DVLs like Vega-Lite and ECharts.}\par

Specifically, in the preparation phase, GRED utilizes a pre-trained text embedding model \cite{reimers2020making,feng2020language} to convert all NLQs and DVQs contained in the nvBench training set into embedding vectors, thus creating an embedding vector repository. Then, ChatGPT is used to generate natural language annotations for each database, creating a collection of annotated database sets. Once ready, for NLQs sent into the text-to-vis system, GRED first uses the pre-trained text embedding model to convert them into embedding vectors and calculates their cosine similarity with the embedding vectors of NLQs in the training set. Then, the top-$K$ most similar NLQs are selected, and their corresponding examples are combined into a generation prompt in descending order of similarity, which is input into ChatGPT to generate the corresponding DVQ, referred to as \text{DVQ}\textsubscript{\text{gen}}. Next, \text{DVQ}\textsubscript{\text{gen}} is converted into embedding vectors, and its cosine similarity with DVQ embedding vectors in the library is calculated. The top-$K$ most similar DVQs are selected to construct a tuning prompt, which is then input into ChatGPT to mimic a similar programming style, resulting in \text{DVQ}\textsubscript{\text{rtn}}. Finally, the database with natural language annotations and \text{DVQ}\textsubscript{\text{rtn}} are combined into a debugging prompt, inputted into ChatGPT to replace inappropriate data schema in \text{DVQ}\textsubscript{\text{rtn}}, obtaining the final \text{DVQ}\textsubscript{\text{dbg}}.\par

Experimental results on nvBench-Rob indicate that GRED significantly surpasses existing text-to-vis models in terms of model robustness. Compared to the current state-of-the-art (SOTA) text-to-vis model RGVisNet, GRED achieves an accuracy improvement of over 20\% on the single-variant test set and over 30\% on the dual-variant test set. These results verify the effectiveness of GRED in enhancing the robustness of text-to-vis models.\par

In a nutshell, the contributions of our work are threefold: 
\begin{itemize}[noitemsep, topsep=0em]
    \item To our knowledge, we are the first to comprehensively study the robustness of the text-to-vis task; We hope this work will inspire more research on improving the robust data visualization models.
    \item We construct nvBench-Rob, the first dedicated dataset to evaluate the robustness of text-to-vis models. We observed significant performance drops of SOTA text-to-vis models on this robustness scenario, revealing that even SOTA models still possess significant potential for further exploration.
    \item We designed a novel framework called GRED, based on RAG technique. This framework effectively addresses the high sensitivity of text-to-vis models to input perturbations and inconsistencies in programming styles. It provides an innovative paradigm for leveraging Large LLMs to tackle robustness issues in the text-to-vis field. 
\end{itemize}

\section{Robustness Dataset: nvBench-Rob}
\subsection{Overview}
We constructed nvBench-Rob benchmark, the first comprehensive robustness evaluation dataset in the field of text-to-vis, through a collaboration between LLMs and humans. Specifically, we utilized LLMs to first modify the original dataset and then manually corrected the modified dataset, which not only saved labor costs but also allowed for diverse language styles and database naming habits within the dataset.\par

In nvBench-Rob, we have meticulously designed three robustness test sets to comprehensively evaluate the models from various perspectives: robustness to NLQs, robustness to table schemas, and robustness to the combination of both.\par
In this section, we will present a detailed overview of our dataset construction method and perform a thorough analysis of the features of nvBench-Rob.

\begin{figure}[t]
    \centering
    \includegraphics[width=0.5\textwidth]{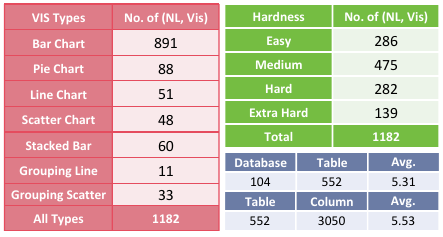}
    \caption{Statistics of the nvBench-Rob Dataset}
    \label{fig:Dataset_statistic}
    \vspace{-15pt}
\end{figure}

\subsection{ChatGPT Modification}
\label{sec:ChatGPT_Modification}
The LLM is a kind of large-scale models trained on a massive corpus, demonstrating outstanding capability in natural language processing (NLP) tasks. ChatGPT is one of these representative models. Through ChatGPT \cite{openai2024gpt4}, we can harness its powerful NLP capability to process the dataset. \par
The existing nvBench dataset usually explicitly mentions table schema (such as column names) and DVQ keywords (e.g., Bin and Group) in the NLQs. This makes it difficult for models trained on this dataset to perform well in scenarios where users have limited knowledge of DV. For instance, users may lack knowledge of table schemas and DVQ syntax (Figure~\ref{fig:Text-to-Vis and Robustness}). During training, the model may only learn the explicit alignment between NLQ, table schemas, and DVQ, rather than truly understanding how to conduct schema linking semantically. This also reflects that nvBench cannot effectively evaluate the robustness of the model.\par
LLMs can be potentially used to address the above issues. With its powerful NLU capability, we can utilize LLMs like ChatGPT to simulate various user interaction behaviors, thereby enhancing the robustness of the text-to-vis dataset.\par
\begin{figure}[htb]
    \centering
    \includegraphics[width=0.5\textwidth]{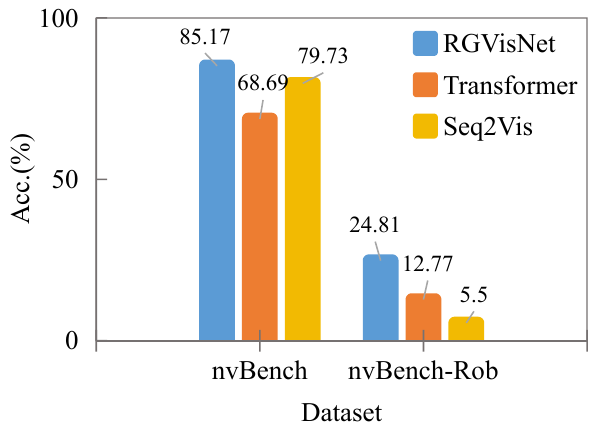}
    \caption{The performance of existing text-to-vis models dramatically \textbf{drops} on the nvBench-Rob datasets.}
    \label{fig:Model_Robustness_a}
    \vspace{-20pt}
\end{figure}
\paragraph{NLQ Reconstruction.} We reconstructed the NLQs in nvBench using ChatGPT, without focusing on explicit mentions of table schema and DVQ keywords within the sentences. Specifically, we replaced most of the nouns in the sentences with synonyms based on the context, aiming to minimize the explicit mention of table schema in the NLQs. With these modifications, we simulated the interaction between a user who is unfamiliar with both the database information and DVQ syntax and the text-to-vis model.\par

\paragraph{Schema Synonymous Substitution.} We attempted to utilize the approach used in MultiSpider \cite{dou2023multispider} by inputting the format ``\textit{table(column)[type]}'' into ChatGPT, with the aim of having it to return a column name with equivalent meaning in that context. However, the results were consistently unsatisfactory. As a result, we refined the method by constructing prompts that included database name, table names, column names, and column types, such as ``\textit{In the `cinema' table `cinema' based on the `filmdom' database, what alternative name could be used for a column with the data type `Text' that conveys a similar meaning to `Movie'? Please return only one English word rather than a sentence.}'' It was empirically demonstrated that this approach yielded superior results. Nevertheless, this method still has several limitations. For instance, in most cases, a table named ``\textit{happy\_hour}'' may have a column named ``\textit{HH\_ID}'', and the model is unaware that ``\textit{HH}'' represents ``\textit{happy\_hour}''. To address these limitations, we made manual modifications.

\subsection{Manual Correction} 
\label{sec:manual_cor}
The output of LLM is characterized by instability. To ensure the efficacy of the dataset, it is necessary for us to undertake manual corrections on the entire dataset. In particular, as mentioned in Section ~\ref{sec:ChatGPT_Modification}, ChatGPT often fails to meet the robustness requirements when performing schema synonym substitution. Hence, we conducted a comprehensive and detailed manual modification of the entire nvBench-Rob dataset. This step constitutes the most critical and valuable aspect of dataset construction.

\subsection{Dataset Analysis}
We randomly divided nvBench into 3 parts according to the ratio of 80/4.5/15.5 in ncNet \cite{luo2021natural}. As a result, we obtained a development set consisting of 1182 pairs of (NL, VIS), involving a total of 104 databases. We performed robustness modifications (i.e. \textit{NLQ reconstruction} and \textit{schema synonymous substitution}) to both the 1182 pairs of (NL, VIS) and schemas in 104 databases. Eventually, three different levels of robustness datasets were obtained: \textbf{nvBench-Rob}\textsubscript{\textit{nlq}}, \textbf{nvBench-Rob}\textsubscript{\textit{schema}}, and \textbf{nvBench-Rob}\textsubscript{\textit{(nlq,schema)}}, corresponding to evaluating robustness modifications only on NLQs, only on table schemas, and on both NLQs and table schemas, respectively. The distribution of visualization chart types and the difficulty level of the DVQs are shown in Figure ~\ref{fig:Dataset_statistic}.

\section{Robustness Analysis of Existing Text-to-Vis Models}
\label{sec:rob_analy}
As shown in Figure~\ref{fig:Model_Robustness_a}, the accuracy of existing text-to-vis models  significantly decreased on the nvBench-Rob test set compared to the nvBench test set. Specifically, on a no-cross-domain split, the previous SOTA text-to-vis model, RGVisNet, achieved an accuracy of 85.17\% on the nvBench test set, and other text-to-vis models also performed satisfactorily. However, even RGVisNet's accuracy dropped to 24.81\% on the nvBench-Rob test set, which comprises both NLQs and data schema variants, marking a 60.36\% decrease compared to its performance on the nvBench test set. This highlights the lack of robustness of the nvBench dataset and the high sensitivity of models trained on it to perturbations in model input.\par
For example, in the nvBench training set, data schemas like column names are explicitly mentioned in the NLQs, such as ``ACC\_Percent,'' enabling text-to-vis models to easily learn the explicit connection between NLQ and data schema. In the nvBench-Rob test set, NLQs no longer explicitly mention database column names, and sentences are reconstructed. Moreover, the column names in the database have been replaced with synonyms, for example, ``ACC\_Percent'' has been replaced with ``percentage\_of\_ACC.'' In these cases, previous text-to-vis models all fail to perform schema linking correctly, with RGVisNet still choosing the same column name "ACC\_Percent" as in the training data, while models like Seq2Vis and Transformer are unable to generate the correct DVQ keywords. For more examples, please refer to Appendix~\ref{more_case}.

\begin{figure*}[htb]
    \centering
    \includegraphics[width=\textwidth]{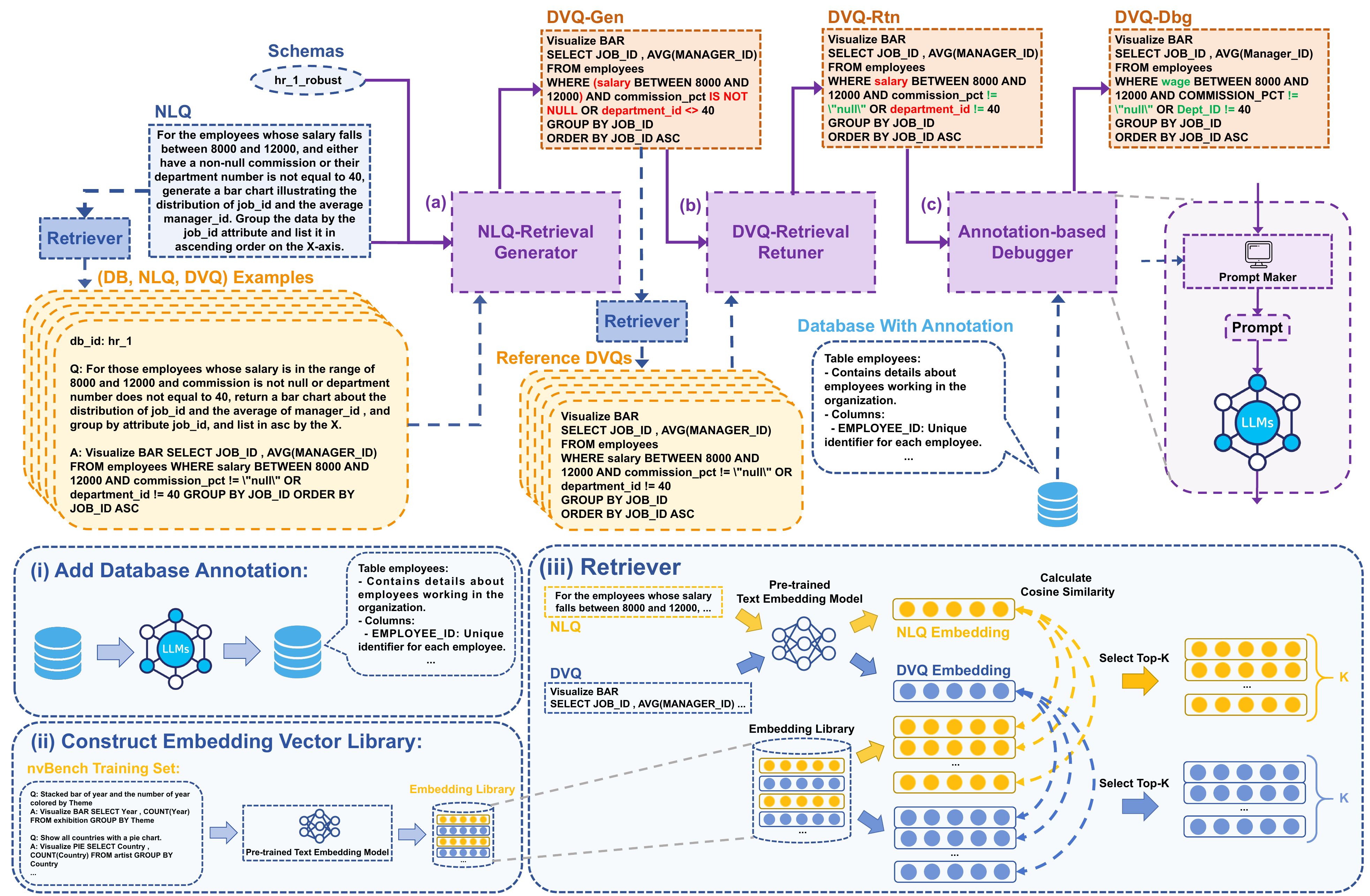}
    \caption{The working pipeline of our proposed GRED method, which includes three steps: (a) Input the NLQ into the Retriever to obtain the top-$K$ (DB, NLQ, Schemas) instances, then input these instances along with the NLQ and Schemas into the \textbf{\textit{NLQ-Retrieval Generator}} to get \textit{DVQ\_Rtn}; (b) Input the DVQ\_Rtn into the Retriever to obtain the top-$K$ DVQs, referred to as Reference DVQs, then input Reference DVQs along with DVQ\_Rtn into the \textbf{\textit{DVQ-Retrieval Retuner}} to get \textit{DVQ\_Rtn}; (c) Input the DVQ\_Rtn and the annotated databases corresponding to Schemas into the \textbf{\textit{Annotation-based Debugger}} to obtain the final result \textit{DVQ\_Dbg}.}
    \label{fig:model}
    \vspace{-0pt}
\end{figure*}

\section{GRED: A Robustness Framework based on Retrieval-Augmented Generation}
To enhance the robustness of text-to-vis models, we propose a novel RAG-based framework, named \textbf{GRED}. This framework comprises three core components: NLQ-Retrieval \underline{G}enerator, DVQ-Retrieval \underline{Re}tuner, and Annotation-based \underline{D}ebugger, aimed at addressing variants of NLQ, differences in programming styles, and changes in data schema, respectively. Before all the main processes of GRED, there are some preparatory works that need to be completed.\par
\subsection{Preparatory Phase}
\label{sec:Preparatory_Works}
The preparatory phase comprises two key steps: the establishment of an embedding vector library and the construction of an annotated database collection. Specifically, for the training set partitioned by nvBench, each NLQ and its corresponding DVQ are input into a pre-trained text embedding model to derive the associated embedding vectors, thereby populating the embedding vector library. The pre-trained text embedding model utilized in this work is the \textit{text-embedding-3-large} model released by OpenAI. Regarding the construction of the annotated database collection, this process entails supplying database information to \textit{GPT-3.5-Turbo} as prompts to generate corresponding NL annotations, which are then stored collectively.\par

\subsection{Pipeline of GRED}
\paragraph{NLQ-Retrieval Generator}
For the NLQs input into the text-to-vis system, GRED first converts them into embedding vectors using the text-embedding-3-large model mentioned in Section~\ref{sec:Preparatory_Works}, and then calculates their cosine similarity with the embedding vectors of natural language questions in the embedding vector library constructed during the preparation. After that, it selects the top-$K$ most similar natural language questions and assembles their corresponding examples into a generation prompt in ascending order of similarity. This prompt is then input into LLM like GPT-3.5-Turbo to generate the corresponding DVQ, referred to as \text{DVQ}\textsubscript{\text{gen}}. It is worth mentioning that sorting in ascending order of similarity means placing examples with high similarity near the asking part of the prompt. For more details, please refer to Appendix~\ref{prompt:generator}. The benefit of this approach is that it allows the LLM to achieve more accurate results based on the examples, thus reducing the model's hallucinations.

\paragraph{DVQ-Retrieval Retuner}

Similar to the retrieval process with NLQ, convert \text{DVQ}\textsubscript{\text{gen}} into embedding vectors, and calculate the cosine similarity with the DVQ embedding vectors in the embedding library constructed in Section~\ref{sec:Preparatory_Works}. Select the top-$K$ most similar DVQs to construct retuning prompts, and then input them into LLM, such as GPT-3.5-Turbo, to mimic similar programming styles, thereby generating \text{DVQ}\textsubscript{\text{rtn}}. The purpose of this step is to perform fine adjustments to the DVQ, such as choosing between ``\textit{IS NOT NULL}'' and ``\textit{!= "null"}''.

\paragraph{Annotation-based Debugger}
The examples in the embedding vector library constructed in Section~\ref{sec:Preparatory_Works} all come from nvBench, which means these examples do not contain data schema variations. This will cause LLMs to experience illusions when encountering data schema variants, resulting in the generation of DVQs with incorrect column names. To tackle this problem, an annotation-based debugger component is introduced. Specifically, this involves combining the database with NL annotations and \text{DVQ}\textsubscript{\text{rtn}} into debugging prompts. Then, inputting them into GPT-3.5-Turbo and asking it to replace the inappropriate column names in \text{DVQ}\textsubscript{\text{rtn}} to obtain the final \text{DVQ}\textsubscript{\text{dbg}}.

In summary, the NLQ-Retrieval Generator ensures that the model's output is structurally similar to the target DVQ. The DVQ-Retrieval Retuner ensures that the model's output closely aligns with the target DVQ in terms of minor programming styles. Lastly, the Annotation-based Debugger guarantees the correctness of the data schema mentioned in the model's output DVQ.

\section{Experiments and Analysis}
In this section, we present the experimental setup and report the evaluation results. Through comparative analysis with other baselines, we demonstrate that our model outperforms baselines in terms of robustness, thus verifying the effectiveness of GRED.
\subsection{Experimental Setup}
\paragraph{Datasets.} We evaluate the robustness of the previous text-to-vis model on the nvBench-Rob test set. The nvBench-Rob test set comprehensively evaluates the model's robustness from three different dimensions: the NLQ single-variant test set, the Data schema single-variant test set, and the dual-variant test set. Therefore, there are three sets of evaluations:
\begin{itemize}[noitemsep,topsep=0em,leftmargin=*]
    \item \textbf{nvBench-Rob}\textsubscript{\textit{nlq}}: a testing set from nvBench-Rob, containing only NLQ variants, is specifically designed to test the robustness of models against NLQ variants.
    \item \textbf{nvBench-Rob}\textsubscript{\textit{schema}}: a testing set from nvBench-Rob, containing only data schema variants, is specifically designed to test the robustness of models against data schema variants.
    \item \textbf{nvBench-Rob}\textsubscript{\textit{(nlq,schema)}}: a testing set from nvBench-Rob, containing both NLQ variants and data schema variants, is specifically designed to test the robustness of models against both NLQ variants and data schema variants.
\end{itemize}\par

\paragraph{Baselines.} We evaluate \textbf{GRED} and previous text-to-vis models on nvBench-Rob including \textbf{Seq2Vis} \cite{10.1145/3448016.3457261}, \textbf{Transformer} \cite{vaswani2017attention}, and \textbf{RGVisNet} \cite{song2022rgvisnet}, which is the previous SOTA model in text-to-vis. We conduct a detailed analysis of the robustness using their performance on nvBench-Rob.

\paragraph{Measurements.} Following \cite{song2022rgvisnet,10.1145/3448016.3457261}, four popular metrics, namely \textit{Vis Accuracy}, \textit{Data Accuracy}, \textit{Axis Accuracy}, and \textit{Overall Accuracy}, are used in our experiment to evaluate the performance. 

\paragraph{Implementation Details.}
For the data preparation phase, specifically for generating NL annotations for each database, the parameters of the \texttt{openai.ChatCompletion.create} method are set as follows: 
\begin{center}
    \texttt{temperature=0.0,\\
    frequency\_penalty=0.0,\\
    presence\_penalty=0.0}
\end{center}
However, during the formal working phase of GRED, the parameters of this function are set as follows: 
\begin{center}
    \texttt{temperature=0.0,\\
    frequency\_penalty=-0.5,\\
    presence\_penalty=-0.5}
\end{center}
In addition, the large language model used in the experimental process is \textit{GPT-3.5-Turbo} and uses the version released by OpenAI on \textit{January 25, 2024}. The hyperparameter $K$, which means the retrieval number of NLQ and DVQ, in the experiment is \textit{10}.
\begin{table}[t!]
    \captionsetup{width=\textwidth}
    \centering
    \resizebox{\linewidth}{!}{
    \begin{tabular}{lcccc}
     & \multicolumn{4}{c}{-\textbf{nvBench-Rob}\textsubscript{\textit{nlq}}}\\
    \cmidrule(lr){2-5}
    \textbf{Model} & \textbf{Vis Acc.} & \textbf{Data Acc.} & \textbf{Axis Acc.} & \textbf{Acc.}\\
    \midrule
    Seq2Vis & 93.91\% & 38.83\% & 42.23\% & 34.52\%\\
    Transformer & 91.62\% & 48.22\% & 49.24\% & 36.04\%\\
    RGVisNet & 96.37\% & 53.04\% & 70.12\% & 45.87\% \\
    GRED (Ours)    & \textbf{97.63}\% & \textbf{61.93}\% & \textbf{88.41}\% & \textbf{59.98}\% \\
    \bottomrule
    \end{tabular}}
    \caption{Results in -\textbf{nvBench-Rob}\textsubscript{\textit{nlq}}}
    \label{tab:results_nlq}
\end{table}

\begin{table}[t!]
    \captionsetup{width=\textwidth}
    \centering
    \resizebox{\linewidth}{!}{
    \begin{tabular}{lcccc}
     & \multicolumn{4}{c}{-\textbf{nvBench-Rob}\textsubscript{\textit{schema}}}\\
    \cmidrule(lr){2-5}
    \textbf{Model} & \textbf{Vis Acc.} & \textbf{Data Acc.} & \textbf{Axis Acc.} & \textbf{Acc.}\\
    \midrule
    Seq2Vis & 96.79\% & 18.02\% & 15.40\% & 14.55\%\\
    Transformer & 92.22\% & 41.88\% & 38.16\% & 29.61\%\\
    RGVisNet & \textbf{98.33\%} & 55.09\% & 60.83\% & 44.91\% \\
    GRED    & 97.72\% & \textbf{65.48\%} & \textbf{85.03\%} & \textbf{61.93\%} \\
    \bottomrule
    \end{tabular}}
    \caption{Results in -\textbf{nvBench-Rob}\textsubscript{\textit{schema}}}
    \label{tab:results_schema}
\end{table}

\begin{table}[t!]
    \captionsetup{width=\textwidth}
    \centering
    \resizebox{\linewidth}{!}{
    \begin{tabular}{lcccc}
     & \multicolumn{4}{c}{-\textbf{nvBench-Rob}\textsubscript{\textit{(nlq,schema)}}}\\
    \cmidrule(lr){2-5}
    \textbf{Model} & \textbf{Vis Acc.} & \textbf{Data Acc.} & \textbf{Axis Acc.} & \textbf{Acc.}\\
    \midrule
    Seq2Vis & 94.16\% & 7.45\% & 7.11\% & 5.50\%\\
    Transformer & 92.13\% & 22.59\% & 18.87\% & 12.77\%\\
    RGVisNet & 96.76\% & 47.04\% & 34.07\% & 24.81\% \\
    GRED    & \textbf{98.14\% }& \textbf{58.48\%} & \textbf{81.52\%} & \textbf{54.85\%} \\
    \bottomrule
    \end{tabular}}
    \caption{Results in -\textbf{nvBench-Rob}\textsubscript{\textit{(nlq,schema)}}}
    \label{tab:results_nlq_schema}
    \vspace{-15pt}
\end{table}

\begin{table*}[th!]
    \captionsetup{width=\textwidth}
    \centering
    \resizebox{\linewidth}{!}{
    \begin{tabular}{l|ccc}
    \textbf{Model} & 
    \textbf{nvBench-Rob}\textsubscript{\textit{nlq}} &
    \textbf{nvBench-Rob}\textsubscript{\textit{schema}} & 
    \textbf{nvBench-Rob}\textsubscript{\textit{(nlq,schema)}}\\
    \midrule
    RGVisNet (SOTA) & 45.87\% & 44.91\% & 24.81\% \\
    GRED (Ours)    & 59.98\% & 61.93\% & \textbf{54.85\%}\\
    $\quad$ - w/o RTN\&DBG & \textbf{62.77\%} & 42.13\% & 36.46\% \\
    $\quad$ - w/o RTN  & 61.08\% & \textbf{62.10\%} & 51.90\%\\
    $\quad$ - w/o DBQ & 61.68\% & 42.47\% & 38.57\%\\
    \bottomrule
    \end{tabular}}
    \caption{Ablation Study Result on nvBench-Rob.}
    \label{tab:ablation_study}
    \vspace{-5pt}
\end{table*}

\begin{table*}[th!]
    \centering
    \resizebox{1.\textwidth}{!}{
    \begin{tabular}{c|m{0.8\linewidth}}
        \toprule
        NLQ & Present the \textbf{\textit{department\_id}} by \textbf{\textit{first name}} in a histogram, with the Y-axis organized in descending order, please. \\

        \cline{1-2}
        Target DVQ & Visualize BAR SELECT \textbf{\textit{Fname}}, \textbf{\textit{Dept\_ID}} FROM employees ORDER BY \textbf{\textit{Dept\_ID}} DESC $\rightarrow$ \textit{Figure~\ref{fig:case_study_Target}}\\

        \cline{1-2}
        Seq2Vis & Visualize BAR SELECT \textcolor{c2}{FIRST\_NAME} , \textcolor{c2}{COUNT (FIRST\_NAME)} FROM \textcolor{c2}{dogs} ORDER BY \textcolor{c2}{COUNT (LAST\_NAME)} DESC $\rightarrow$ \textit{Figure~\ref{fig:Seq2Vis_case}}\\
        \cline{1-2}
        Transformer & Visualize BAR SELECT \textcolor{c2}{FIRST\_NAME} , \textcolor{c2}{DEPARTMENT\_ID} FROM employees ORDER BY \textcolor{c2}{DEPARTMENT\_ID} DESC $\rightarrow$ \textit{Figure~\ref{fig:Transformer_case}}\\
        \cline{1-2}
        RGVisNet & Visualize BAR SELECT \textcolor{c2}{FIRST\_NAME} , \textcolor{c2}{DEPARTMENT\_ID} FROM employees ORDER BY \textcolor{c2}{DEPARTMENT\_ID} DESC $\rightarrow$ \textit{Figure~\ref{fig:RGVisNet_case}} \\
        \cline{1-2}
        GRED & Visualize BAR SELECT \textcolor{c1}{\textbf{Fname}} , \textcolor{c1}{\textbf{Dept\_ID}} FROM employees ORDER BY \textcolor{c1}{\textbf{Dept\_ID}} DESC $\rightarrow$ \textit{Figure~\ref{fig:GRED_case}} \\
        \bottomrule

    \end{tabular}
    }
        \begin{subfigure}[t]{0.30\textwidth}
            
            \includegraphics[width=\textwidth]{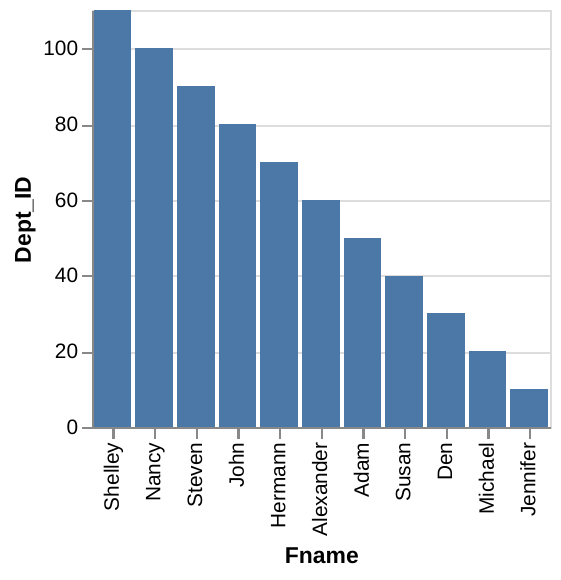}
            \caption{\\Target}
            \label{fig:case_study_Target}
        \end{subfigure}
        \hfill
        \begin{subfigure}[t]{0.12\textwidth}
           
            \includegraphics[width=\textwidth]{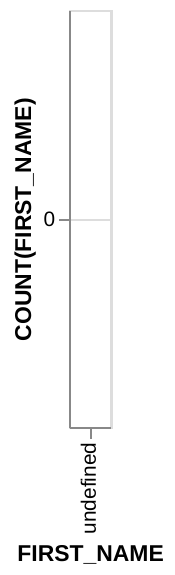}
             \caption{\\Seq2Vis \textcolor{c2}{\ding{56}}}
            \label{fig:Seq2Vis_case}
        \end{subfigure}
        \hfill
        \begin{subfigure}[t]{0.12\textwidth}
             
            \includegraphics[width=\textwidth]{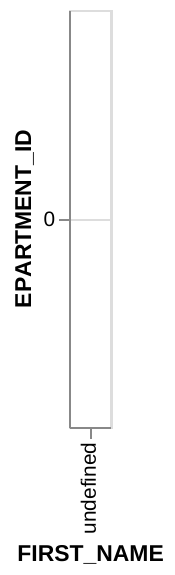}
           \caption{\\Transformer \textcolor{c2}{\ding{56}}}
            \label{fig:Transformer_case}
        \end{subfigure}
        \hfill
        \begin{subfigure}[t]{0.12\textwidth}
            
            \includegraphics[width=\textwidth]{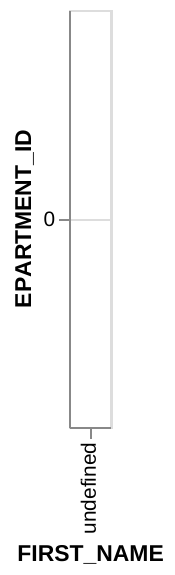}
            \caption{\\RGVisNet \textcolor{c2}{\ding{56}}}
            \label{fig:RGVisNet_case}
        \end{subfigure}
        \hfill
        \begin{subfigure}[t]{0.30\textwidth}
            \centering
            
            \includegraphics[width=\textwidth]{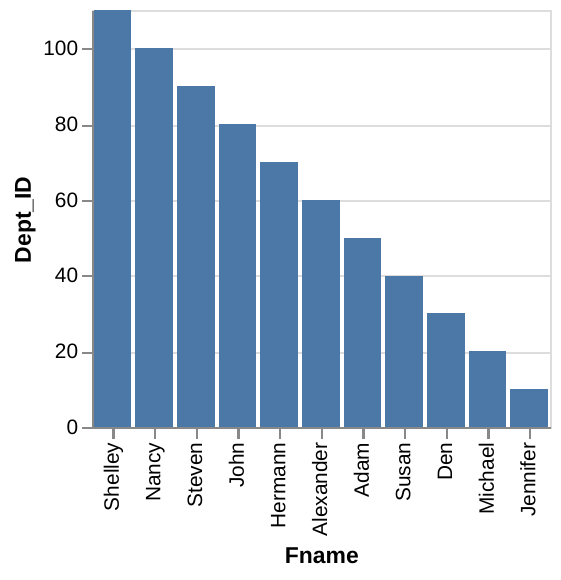}
            \caption{\\GRED \textcolor{c1}{\ding{52}}}
            \label{fig:GRED_case}
        \end{subfigure}
    \caption{Case Study. DVQs generated by other baselines like RGVisNet and GRED, together with their corresponding visualization charts (Errors are marked with \textcolor{red}{red} colors).}
    
    \label{tab:case_study}
    \vspace{-10pt}
\end{table*}

\subsection{Experiment Result}
As shown in Figure~\ref{fig:Model_Robustness_a}, previous text-to-vis models have achieved satisfactory performance on the nvBench test set. Even the simplest model, Seq2Vis, can easily achieve high precision. However, when the model input is perturbed, even the state-of-the-art(SOTA) model RGVisNet experiences a significant drop in accuracy.\par

In order to comprehensively assess the robustness of the models, we tested the models trained on nvBench with three test sets from nvBench-Rob. The results are presented in Table~\ref{tab:results_nlq}, Table~\ref{tab:results_schema} and Table~\ref{tab:results_nlq_schema}. The SOTA text-to-vis model, RGVisNet, experienced a significant decline of 39.3\% (85.17\% vs. 45.87\%) or 40.26\% (85.17\% vs. 44.91\%) in accuracy on test sets with a single variation. The most notable difference was observed in nvBench-Rob, where the accuracy dropped by 60\% (85.17\% vs. 24.81\%) compared to the original nvBench test set. Meanwhile, GRED demonstrated impressively high accuracy across the three test sets of nvBench-Rob, with an improvement of 16\% (61.68\% vs. 45.87\%) on \text{nvBench-Rob}\textsubscript{\text{nlq}}, 18.5\% (63.45\% vs. 44.91\%) on \text{nvBench-Rob}\textsubscript{\text{schema}}, and 32\% (57.19\% vs. 24.81\%) on the most challenging \text{nvBench-Rob}\textsubscript{\text{(nlq,schema)}} test set. Such results indicate that GRED has a strong ability to resist interference with model inputs and demonstrate its excellent robustness.

\subsection{Ablation Study}
In this section, we conduct ablation studies to demonstrate the effectiveness and contribution of each design component in GRED. Specifically, we first evaluate GRED with all components included. Then, we remove some components of GRED to assess its performance with the following configurations: (i) utilizing only NLQ-Retrieval Generator without DVQ-Retrieval Retuner and Annotation-based Debugger \textbf{(w/o RTN\&DBG)}; (ii) removing our Annotation-based Debugger \textbf{(w/o DBG)}; (iii) removing the DVQ-Retrieval Retuner \textbf{(w/o RTN)}. \par
The ablation study results shown in Table~\ref{tab:ablation_study} confirm the importance of the three components designed in our proposed model. We observed that the NLQ-Retrieval Generator plays a crucial role in countering input perturbations caused by natural language variants, while the Annotation-based Debugger plays a key role in countering input perturbations caused by data schema variations. This is because they significantly improve the model's performance in their respective variant-specific test sets. The DVQ-Retrieval Retuner is also found to be very important since it helps LLM adjust the generated DVQ style to better match the dataset's style, thereby reducing errors in programming style and achieving higher accuracy. Therefore, these three components all contribute to the model's robustness.

\subsection{Case Study}
Table~\ref{tab:case_study} presents a case study illustrating the DVQ generated by GRED and previous SOTA model RGVisNet. The charts generated by these models are also shown in Table~\ref{tab:case_study}. As illustrated by Table~\ref{tab:case_study}, Seq2Vis generate incorrect column names and aggregation keywords on y-axis, resulting in no chart being shown in Table~\ref{fig:Seq2Vis_case}. RGVisNet and Transformer generate DVQs with the correct aggregation keywords. However, due to the lack of robustness to perturbations in model inputs, both RGVisNet and Transformer fail to accurately generate the column names "Fname" and "Dept\_ID". Instead, they retain the column names "FIRST\_NAME" and "DEPARTMENT\_ID" from the training set, which also results in no chart being produced in Table~\ref{fig:RGVisNet_case} and Table~\ref{fig:Transformer_case}. Unlike the aforementioned models, GRED is capable of not only generating a structure that is identical to the target query but also producing the correct column names, thereby resulting in the accurate charts as shown in Table~\ref{fig:GRED_case}.

\section{Related Work}
\vspace{-5pt}
\paragraph{Automatic Data Visualization.}
Recent years, there has been significant growth in the adoption of Data Visualization (DV) in the fields of natural language processing \cite{ge2022chinese}, data mining \cite{song2022rgvisnet,qian2021learning,ho2002visualization,fayyad2002information}, and database community \cite{tang2022sevi,hanrahan2006vizql,10.1145/3448016.3457261,vartak2017towards,luo2018deepeye}. Various advanced techniques have been developed to simplify the use of DV. We will cover two typical approaches: text-to-vis and DV Recommendation. \par
Text-to-vis primarily aims to convert a natural language question (NLQ) into its equivalent DV, simplifying the process for non-specialist users. The prevailing approach treats this conversion similarly to machine translation, engaging deep neural networks to map NLQs to DVs. For instance, Cui \textit{et al.} introduced the concept of text-to-viz, employing rule-based systems to convert text into infographics \cite{cui2019text}. In contrast, Draco-Learn \cite{moritz2018formalizing} views visualization design as a selection from constraints, adapting weights based on experimental data. Data2Vis \cite{dibia2019data2vis} translates data into visualization specifications using machine translation frameworks. NL4DV \cite{narechania2020nl4dv} offers a Python-based toolkit that provides high-level functions to aid the construction of DV with natural language interfaces. Luo \textit{et al.} also delineated a methodology for synthesizing the NLQ-DV dataset, known as nvBench, predicated upon the renowned NL2SQL benchmark, Spider \cite{yu-etal-2018-spider}. A Seq2Seq model was subsequently trained on this benchmark \cite{10.1145/3448016.3457261}, corroborating the viability of engendering DV queries from NLQs. RGVisNet \cite{song2022rgvisnet} represents another seminal study in which a DNN-based approach is employed to transform NLQ into DV. On the other hand, automated DV recommendation systems output probable DVs from datasets without any NLQ involvement. DataEye \cite{luo2018deepeye} simplifies this problem into recognition, ranking, and selection steps. \newcite{qian2021learning} introduced an end-to-end learning-based approach for constructing DVs from extensive datasets.\par
Despite the abundance of models in text-to-vis, the robustness of these models remains underexplored. We introduce nvBench-Rob, the first dataset designed to comprehensively evaluate the robustness of existing text-to-vis models. 
Furthermore, we propose a novel RAG-based framework called GRED, designed to address perturbations in the model’s input from three aspects: NLQ variants, programming style differences, and data schema variants. With the benchmark and the method proposed in this paper, nvBench-Rob would become a popular dataset for evaluating the robustness of text-to-vis models and inspire further research in the \textit{NLP for Visualization} direction.
\paragraph{Robustness in NLP.}
The robustness of a model is a crucial evaluation criterion for its deployment in real-life scenarios. In the field of NLP, there have been numerous studies on model robustness. Some studies have investigated the influence of model inputs on robustness. For example, \newcite{chen-etal-2022-rationalization} examined how NLP models behave under various 'AddText' attacks and evaluates their performance across various NLP tasks.\newcite{yan-etal-2022-robustness} studied the robustness of NLP models' reading comprehension when faced with entity renaming. Furthermore, \newcite{hendrycks2019pretraining,hendrycks2020pretrained} showed that pre-training can enhance a model's robustness against label noise, class imbalance, and out-of-distribution detection. Other studies have proposed domain-specific optimization methods for model robustness. For example, \newcite{zhao-etal-2023-robut} introduced a framework for generating table adversarial training examples using LLMs for enhancing the robustness of TableQA models. In cross-domain question-answering, \newcite{chen-etal-2023-improving-robustness} improved robustness through a dual-data augmentation approach. Besides, some studies have introduced evaluation metrics to evaluate model robustness across various domains. For example, \newcite{wang-etal-2023-recode} proposed over 30 natural transformations specific to code, such as docstrings, function names, syntax, and format, to evaluate the robustness of code generation models. In another study, \newcite{zhao-etal-2023-robut} introduced a benchmark named ROBUT, which incorporates human-annotated perturbations in table headers, table content, and questions to evaluate the robustness of Table QA models. A comprehensive survey on Robustness in NLP can be found in \cite{wang-etal-2022-measure}. \par
In our research, to thoroughly assess the robustness of text-to-vis models, we introduce a robustness evaluation dataset called nvBench-Rob. Through collaboration between a LLM and human annotators, this dataset not only removes explicit mentions of column names in NLQs but also integrates a variety of language styles in the NLQs. Furthermore, it encompasses a variety of naming conventions for the table schemas in the dataset, thus creating an exceptionally robust evaluation dataset.\par

\paragraph{RAG in NLP}
Retrieval-Augmented Generation (RAG) technology has become the primary method to fully utilize the capabilities of LLMs in downstream tasks~\cite{kim-etal-2023-tree,ma-etal-2023-query,long-etal-2023-adapt,pozzobon-etal-2023-goodtriever,yu-etal-2023-retrieval,shao-etal-2023-enhancing,mavi-etal-2023-retrieval}. It has achieved notable results in various tasks such as open-domain QA~\cite{izacard2021leveraging,trivedi-etal-2023-interleaving,li-etal-2023-large}, dialogue~\cite{cai-etal-2019-skeleton,cai-etal-2019-retrieval,peng2023check}, domain-specific question answering~\cite{cui2023chatlaw} and code generation~\cite{zhou2023docprompting}.

We introduced a RAG-based framework called GRED, which effectively addresses this issue by breaking down the visualization query generation process into subprocess, progressively approximating the ultimate goal. This approach fully utilizes the role of outdated external knowledge. We are the first to validate the effectiveness of the RAG technique in the robust text-to-vis scenario.

\section{Conclusion}
Robustness is a crucial factor for evaluating model performance. In this study, we introduce the first comprehensive robustness benchmark, nvBench-Rob, for evaluating the robustness of text-to-vis models. Then, we found that the performance of existing text-to-vis models is not satisfactory on the robustness scenario. Finally, we propose a novel framework named GRED based on the RAG-techniques using LLMs, which addresses challenges posed by NLQ variations, programming style differences, and data schema variations through three components: NLQ-Retrieval Generator, DVQ-Retrieval Retuner, and Annotation-based Debugger. Our experiments reveal the inherent difficulty of developing robust text-to-vis models, and simultaneously demonstrate the effectiveness of GRED through extensive empirical validation.

\section*{Acknowledgements}
We thanks the reviewers for their valuable comments.

\bibliography{custom}

\newpage
\onecolumn
\appendix

\section{Detailed Definitions of the Evaluation Metrics}
\begin{itemize}[topsep=0em]
    \item \underline{Overall Accuracy}: This metric measures exact matches between the predicted DV query and the target DV query. The accuracy calculation formula is:\\
    \makecell[c]{\texttt{Acc. = \text{N}\textsubscript{c} / N}}\\
    where \text{N}\textsubscript{c} represents the number of the matched DV queries and \text{N} represents the size of the test set. This metric directly reflects the comprehensive performance of the model.
    \item \underline{Vis Accuracy}: Each DVQ consists of three types of components: the DV chart type, the x/y-axis, and the data transformation. This evaluation metric reflects the matches between the generated DVQ and the target DVQ in terms of the type of DV chart. The accuracy calculation formula is:\\
    \makecell[c]{\texttt{Vis Acc. = \text{N}\textsubscript{Vis} / N}}\\
    Where \text{N}\textsubscript{Vis} represents the number of DV chart types in the generated DVQs that match the DV chart types in the target DVQs.
    \item \underline{Axis Accuracy}: This evaluation metric calculates the matches of the x/y axis components between the generated DVQs and the real DVQs. The accuracy calculation formula is:\\
    \makecell[c]{\texttt{Axis Acc. = \text{N}\textsubscript{Axis} / N}}\\
    where \text{N}\textsubscript{Axis} represents the number of x/y-axis components in the generated DVQs that match the x/y-axis components in the target DVQs.
    \item \underline{Data Accuracy}: Similarly, this measurement reflects the matches of the data transformation components between the generated DVQs and the target DVQs. The accuracy calculation formula is:\\
    \makecell[c]{\texttt{Data Acc. = \text{N}\textsubscript{Data} / N}}\\
    where \text{N}\textsubscript{Data} represents the number of data transformation components in the generated DVQs that match the Ddata transformation components in the target DVQs.
\end{itemize}

\newpage
\section{Robustness Analysis Cases}
\label{more_case}

\begin{figure*}[htbp!]
    \centering
    \begin{subfigure}[b]{0.8\textwidth}
        \includegraphics[width=\textwidth]{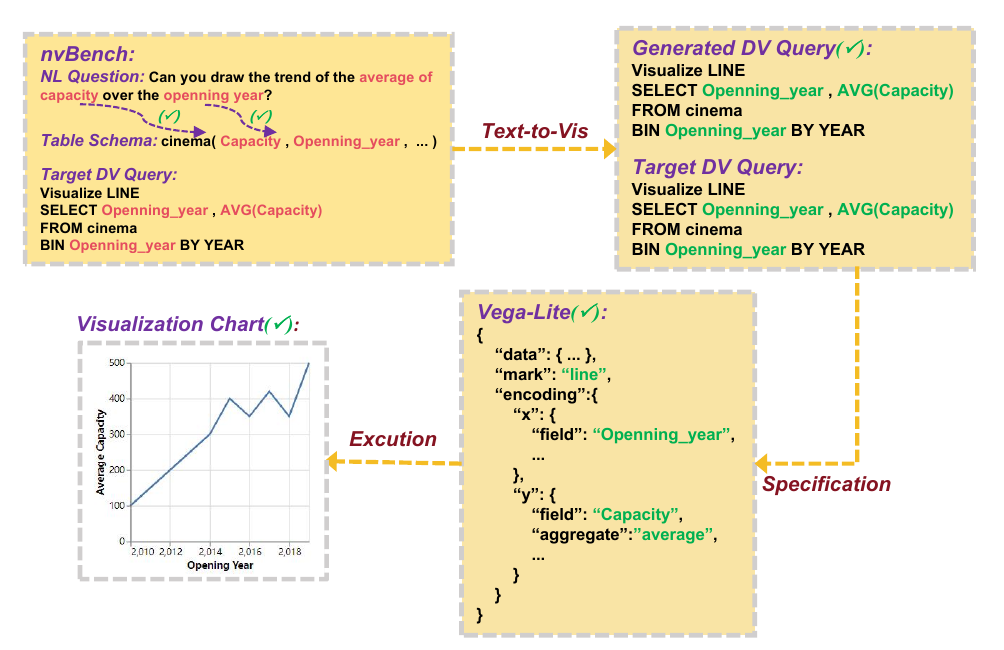}
        \caption{Correct Case in the original Text-to-Vis testing set}
        \label{fig:Case_correct}
    \end{subfigure}
    \begin{subfigure}[b]{0.8\textwidth}
        \includegraphics[width=\textwidth]{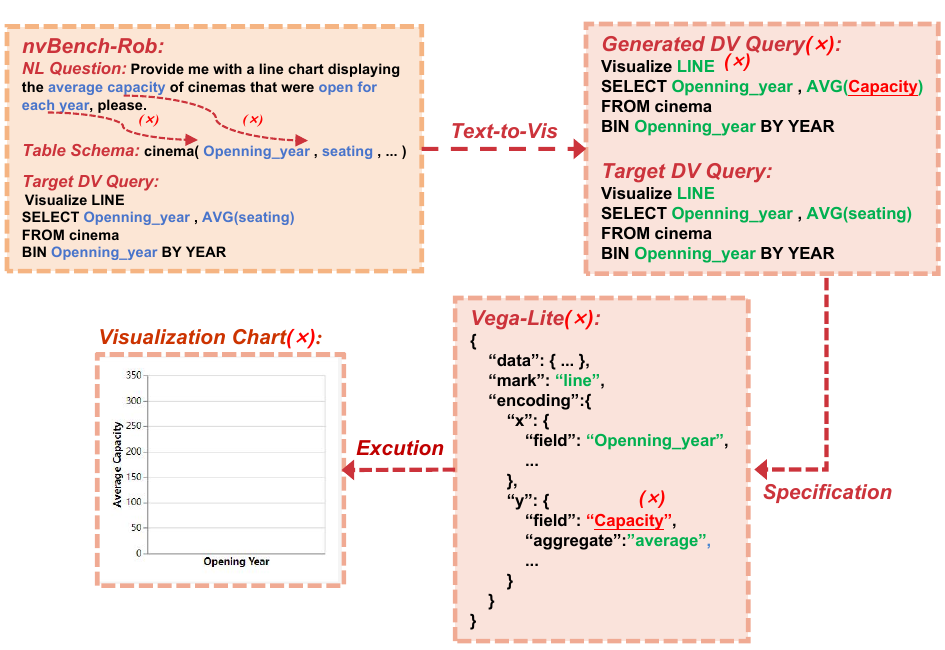}
        \caption{Failure Case of existing Text-to-Vis Models on the Robustness scenario.}
        \label{fig:Case_wrong}
    \end{subfigure}
    \caption{Robustness Analysis Cases}
    \label{fig:Robustness_Analysis_Cases}
\end{figure*}
Figures~\ref{fig:Case_correct} and Figures~\ref{fig:Case_wrong} show examples of previous text-to-vis models successfully generating accurate data visualizations on the original nvBench test set, as well as instances where they fail to produce the final data visualizations due to the addition of NLQ variants and data schema variants. It is not difficult to observe that when the explicit alignment between NLQ and data schema is eliminated, previous text-to-vis models are unable to correctly perform schema linking, even when the data schema has the same meaning as the data schemas in the original training set.

\newpage
\section{Prompts}
\subsection{Database Annotation Generator}
\label{prompt:database_ann}
\noindent \textbf{Role:} \textcolor{teal}{SYSTEM} \\
\textbf{Content:} \\
You are a data mining engineer with ten years of experience in data visualization.\\

\noindent \textbf{Role:} \textcolor{red}{USER} \\
\textbf{Content:} \\
\#\#\#\# Please generate detailed natural language annotations to the following database schemas.\\
\#\#\# Database Schemas:\\
\# Table departments, columns = [ * , Dept\_ID , Dept\_NAME , Manager\_ID , Location\_ID ]\\
\# Table job\_history , columns = [ * , employee\_id , START\_DATE , END\_DATE , JOB\_ID , Dept\_ID ]\\
\# Table jobs , columns = [ * , JOB\_ID , JOB\_TITLE , minimum\_salary , maximum\_salary ]\\
\# Foreign\_keys = [ job\_history.JOB\_ID = jobs.JOB\_ID , job\_history.Dept\_ID = departments.Dept\_ID ]\\\\
\#\#\# Natural Language Annotations:\\
A:\\
Table departments:\\
- Stores data related to different departments within an organization.\\
- Columns:\\
  - Dept\_ID: Unique identifier for each department.\\
  - Dept\_NAME: Name of the department.\\
  - Manager\_ID: Identifier of the manager of the department.\\
  - Location\_ID: Identifier of the location where the department is situated.\\\\
Table job\_history:\\
- Stores historical data of job changes for employees.\\
- Columns:\\
  - employee\_id: Identifier of the employee.\\
  - START\_DATE: Start date of the job role.
  - END\_DATE: End date of the job role.\\
  - JOB\_ID: Identifier of the job role.\\
  - Dept\_ID: Identifier of the department during the job role.\\\\
Table jobs:\\
- Contains information about different job roles.\\
- Columns:\\
  - JOB\_ID: Unique identifier for each job role.\\
  - JOB\_TITLE: Title of the job role.\\
  - minimum\_salary: Minimum salary for the job role.\\
  - maximum\_salary: Maximum salary for the job role.\\\\
Foreign Keys:\\
- job\_history.JOB\_ID references jobs.JOB\_ID, linking job history to specific job roles.\\
- job\_history.Dept\_ID references departments.Dept\_ID, connecting job history to departments.\\\\
\#\#\# Database Schemas:\\
\textit{*****[new Database Schemas]*****}\\\\
\#\#\# Natural Language Annotations:\\
A:

\newpage
\subsection{NLQ-Retrieval Generator Prompt}
\label{prompt:generator}
\noindent \textbf{Role:} \textcolor{teal}{SYSTEM} \\
\textbf{Content:} \\
Please follow the syntax in the examples instead of SQL syntax.\\

\noindent \textbf{Role:} \textcolor{red}{USER} \\
\textbf{Content:} \\
\#\#\#\# Given Natural Language Questions, Generate DVQs based on their correspoding Database Schemas.\\\\
\textit{*****[Top-$K-1$ Examples]*****}\\\\
\#\#\# Database Schemas:\\
\# Table Has\_Pet, columns = [ * , StuID , PetID ]\\
\# Table Pets, columns = [ * , PetID , PetType , pet\_age , weight ]\\
\# Table Student, columns = [ * , StuID , LName , Fname , Age , Sex , Major , Advisor , city\_code ]\\
\# Foreign\_keys = [ Has\_Pet.StuID = Student.StuID , Has\_Pet.PetID = Pets.PetID ]\\
\#\\
\#\#\# Chart Type: [ BAR , PIE , LINE , SCATTER ]\\
\#\#\# Natural Language Question:\\
\# “Find the id and weight of all pets whose age is older than 1 Visualize by bar chart, sort by the Y-axis from high to low.”\\
\#\#\# Data Visualization Query:\\
A: Visualize BAR SELECT PetID , weight FROM pets WHERE pet\_age > 1 ORDER BY weight DESC\\\\
\#\#\# Database Schemas:\\
\# Table Has\_Pet, columns = [ * , StuID , PetID ]\\
\# Table Pets, columns = [ * , PetID , PetType , pet\_age , weight ]\\
\# Table Student, columns = [ * , StuID , LName , Fname , Age , Sex , Major , Advisor , city\_code ]\\
\# Foreign\_keys = [ Has\_Pet.StuID = Student.StuID , Has\_Pet.PetID = Pets.PetID ]\\
\#\\
\#\#\# Chart Type: [ BAR , PIE , LINE , SCATTER ]\\
\#\#\# Natural Language Question:\\
\# “Find the id and weight of all pets whose age is older than 1 Visualize by bar chart, sort in descending by the names.”\\
\#\#\# Data Visualization Query:

\newpage
\subsection{DVQ-Retrieval Retuner}
\label{prompt:Retuner}
% \subsection{Ask1}
\noindent \textbf{Role:} \textcolor{teal}{SYSTEM} \\
\textbf{Content:} \\
The Reference Data Visualization Queries(DVQs) all comply with the syntax of DVQ. Please follow the syntax of the referenced DVQ to modify the Original DVQ.

\noindent \textbf{Role:} \textcolor{red}{USER} \\
\textbf{Content:} \\
\#\#\# Reference DVQs:\\
\textit{*****[Top-$K-1$ DVQs]*****}\\
10 - Visualize BAR SELECT JOB\_ID , SUM(DEPARTMENT\_ID) FROM employees WHERE first\_name LIKE '\%D\%' OR first\_name LIKE '\%S\%' GROUP BY JOB\_ID ORDER BY SUM(DEPARTMEN)\\\\
\#\#\#\# Given the Reference DVQs, please modify the Original DVQ to mimic the style of the Reference DVQs.\\
\#\#\#\# NOTE: Do not Modify the column name in Original DVQ. Especially do not Modify the column names in the ORDER clause!\\
\#\#\# Original DVQ:\\
\# Visualize BAR SELECT JOB\_ID , COUNT(DISTINCT JOB\_ID) FROM employees WHERE DEPARTMENT\_ID = (SELECT DEPARTMENT\_ID FROM departments WHERE DEPARTMENT\_NAME = Finance)\\
A: Let’s think step by step! \\

\noindent \textbf{Role:} \textcolor{blue}{ASSISTANT} \\
\textbf{Content:} \\
\#\#\# Modified DVQ:\\
\# Visualize BAR SELECT JOB\_ID , COUNT(JOB\_ID) FROM employees AS T1 JOIN departments AS T2 ON T1.DEPARTMENT\_ID = T2.DEPARTMENT\_ID WHERE T2.DEPARTMENT\_NAME = 'Finance' GROUP BY JOB\_ID\\

\noindent \textbf{Role:} \textcolor{red}{USER} \\
\textbf{Content:} \\
\#\#\# Reference DVQs:\\
\textit{*****[Top-$K$ DVQs]*****}\\\\
\#\#\#\# Given the Reference DVQs, please modify the Original DVQ to mimic the style of the Reference DVQs.\\
\#\#\#\# NOTE: Do not Modify the column name in Original DVQ. Especially do not Modify the column names in the ORDER clause!\\
\#\#\# Original DVQ:\\
\# \textit{[Original DVQ]}\\
A: Let’s think step by step!

\newpage
\subsection{Annotation-based Debugger}
\noindent \textbf{Role:} \textcolor{teal}{SYSTEM} \\
\textbf{Content:} \\
\#\#\#\# NOTE: Don’t replace column names in Original DVQ that already exist in the database schemas, especially column names in GROUP BY Clause!\\

\noindent \textbf{Role:} \textcolor{red}{USER} \\
\textbf{Content:} \\
\#\#\#\# Please generate detailed natural language annotations to the following database schemas.\\
\#\#\# Database Schemas:\\
\# Table departments, columns = [ * , Dept\_ID , Dept\_NAME , Manager\_ID , Location\_ID ]\\
\# Table job\_history , columns = [ * , employee\_id , START\_DATE , END\_DATE , JOB\_ID , Dept\_ID ]\\
\# Table jobs , columns = [ * , JOB\_ID , JOB\_TITLE , minimum\_salary , maximum\_salary ]\\
\# Foreign\_keys = [ job\_history.JOB\_ID = jobs.JOB\_ID , job\_history.Dept\_ID = departments.Dept\_ID ]\\\\
\#\#\# Natural Language Annotations:\\
A:\\
Table departments:\\
- Stores data related to different departments within an organization.\\
- Columns:\\
  - Dept\_ID: Unique identifier for each department.\\
  - Dept\_NAME: Name of the department.\\
  - Manager\_ID: Identifier of the manager of the department.\\
  - Location\_ID: Identifier of the location where the department is situated.\\\\
Table job\_history:\\
- Stores historical data of job changes for employees.\\
- Columns:\\
  - employee\_id: Identifier of the employee.\\
  - START\_DATE: Start date of the job role.
  - END\_DATE: End date of the job role.\\
  - JOB\_ID: Identifier of the job role.\\
  - Dept\_ID: Identifier of the department during the job role.\\\\
Table jobs:\\
- Contains information about different job roles.\\
- Columns:\\
  - JOB\_ID: Unique identifier for each job role.\\
  - JOB\_TITLE: Title of the job role.\\
  - minimum\_salary: Minimum salary for the job role.\\
  - maximum\_salary: Maximum salary for the job role.\\\\
Foreign Keys:\\
- job\_history.JOB\_ID references jobs.JOB\_ID, linking job history to specific job roles.\\
- job\_history.Dept\_ID references departments.Dept\_ID, connecting job history to departments.\\\\
\#\#\#\# Given Database Schemas and their corresponding Natural Language Annotations, Please replace the column names in the Data Visualization Query(DVQ, a new Programming Language abstracted from Vega-Zero) that do not exist in the database.\\
\#\#\#\# NOTE: Don’t replace column names in Original DVQ that already exist in the database schemas, especially column names in GROUP BY Clause!\\
\#\#\# Original DVQ:\\
\# Visualize BAR SELECT jobid , COUNT(jobid) FROM employees AS T1 JOIN departments AS T2 ON T1.DEPARTMENT\_ID = T2.DEPARTMENT\_ID WHERE T2.DEPARTMENT\_NAME = 'Finance' GROUP BY FIRST\_NAME\\
A: Let’s think step by step! \\
\\

\noindent \textbf{Role:} \textcolor{blue}{ASSISTANT} \\
\textbf{Content:} \\
\#\#\# Revised DVQ:\\
\# Visualize BAR SELECT JOB\_ID , COUNT(JOB\_ID) FROM employees AS T1 JOIN departments AS T2 ON T1.Dept\_ID = T2.Dept\_ID WHERE T2.Dept\_NAME = 'Finance' GROUP BY FIRST\_NAME

\noindent \textbf{Role:} \textcolor{teal}{SYSTEM} \\
\textbf{Content:} \\
\#\#\#\# NOTE: Don’t replace column names in Original DVQ that already exist in the database schemas, especially column names in GROUP BY Clause!\\

\noindent \textbf{Role:} \textcolor{red}{USER} \\
\textbf{Content:} \\
\#\#\#\# Please generate detailed natural language annotations to the following database schemas.\\
\#\#\# Database Schemas:\\
\textit{*****[Database Schemas]*****}\\\\
\#\#\# Natural Language Annotations:\\
\textit{*****[NL Annotation of Database Schemas]*****}\\\\
\#\#\#\# Given Database Schemas and their corresponding Natural Language Annotations, Please replace the column names in the Data Visualization Query(DVQ, a new Programming Language abstracted from Vega-Zero) that do not exist in the database.\\
\#\#\#\# NOTE: Don’t replace column names in Original DVQ that already exist in the database schemas, especially column names in GROUP BY Clause!\\
\#\#\# Original DVQ:\\
\# \textit{[Original DVQ]}\\
A: Let’s think step by step! 
% This is a section in the appendix.

\end{document}